\documentclass{article}

\usepackage{arxiv}
\usepackage{amsmath,amssymb} 
\usepackage[utf8]{inputenc} 
\usepackage[T1]{fontenc}    
\usepackage{hyperref}       
\usepackage{url}            
\usepackage{booktabs}       
\usepackage{amsfonts}       
\usepackage{nicefrac}       
\usepackage{microtype}      
\usepackage{cleveref}       
\usepackage{lipsum}         
\usepackage{graphicx}
\usepackage[numbers]{natbib}
\usepackage{doi}
\usepackage{orcidlink}

\usepackage{tabularx} 
\usepackage{booktabs}
\newcolumntype{Y}{>{\centering\arraybackslash}X}
\newcommand{\eg}{\textit{e}.\textit{g}.}
\newcommand{\ie}{\textit{i}.\textit{e}.}
\newcommand{\Arch}[2]{$\binom{#1}{#2}$}
\title{Is Appearance Free Action Recognition Possible?}

\author{
Filip Ilic \orcidlink{0000-0002-2102-4816} \\
TU Graz, Austria \\
\texttt{filip.ilic@tugraz.at} \\
\And
Thomas Pock \orcidlink{0000-0001-6120-1058} \\
TU Graz, Austria \\
\texttt{pock@icg.tugraz.at} \\
\And
Richard P. Wildes \orcidlink{0000-0003-3433-1329} \\
York University, Canada \\
\texttt{wildes@cse.yorku.ca} \\

}

\renewcommand{\headeright}{}
\renewcommand{\undertitle}{}
\renewcommand{\shorttitle}{Is Appearance Free Action Recognition Possible?}

\hypersetup{
pdftitle={Is Appearance Free Action Recognition Possible?},
pdfsubject={cs.CV, cs.ML},
pdfauthor={Filip Ilic, Thomas Pock, Richard P. Wildes},
	pdfkeywords={Action Recognition, Action Recognition Dataset, Deep Learning, Static and Dynamic Video Representation},
}

\begin{document}
\maketitle

\begin{abstract}
	Intuition might suggest that motion and dynamic information are key to video-based action recognition. In contrast, there is evidence that state-of-the-art deep-learning video understanding architectures are biased toward static information available in single frames. Presently, a methodology and corresponding dataset to isolate the effects of dynamic information in video are missing.
	Their absence makes it difficult to understand how well contemporary architectures capitalize on dynamic vs. static information. We respond with a novel Appearance Free Dataset (AFD) for action recognition. AFD is devoid of static information relevant to action recognition in a single frame. Modeling of the dynamics is necessary for solving the task, as the action is only apparent through consideration of the temporal dimension.
	We evaluated 11 contemporary action recognition architectures on AFD as well as its related RGB video. Our results show a notable decrease in performance for all architectures on AFD compared to RGB. We also conducted a complimentary study with humans that shows their recognition accuracy on AFD and RGB  is very similar and much better than the evaluated architectures on AFD. Our results motivate a novel architecture that revives explicit recovery of optical flow, within a contemporary design for best performance on AFD and RGB.
\end{abstract}

\keywords{Action Recognition, Action Recognition Dataset, Deep Learning, Static and Dynamic Video Representation}

\begin{figure}[h]
	\begin{center}
		\begin{tabular}{lr}
			\fbox{\includegraphics[width=0.5\linewidth]{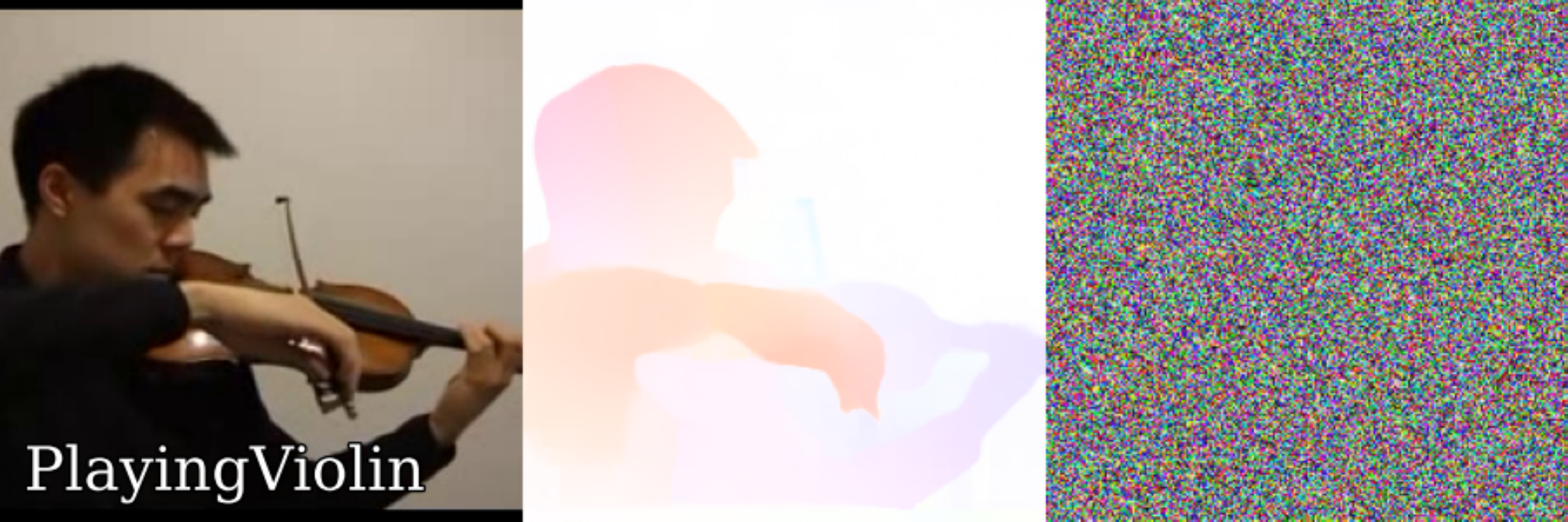}} & \fbox{\includegraphics[width=0.5\linewidth]{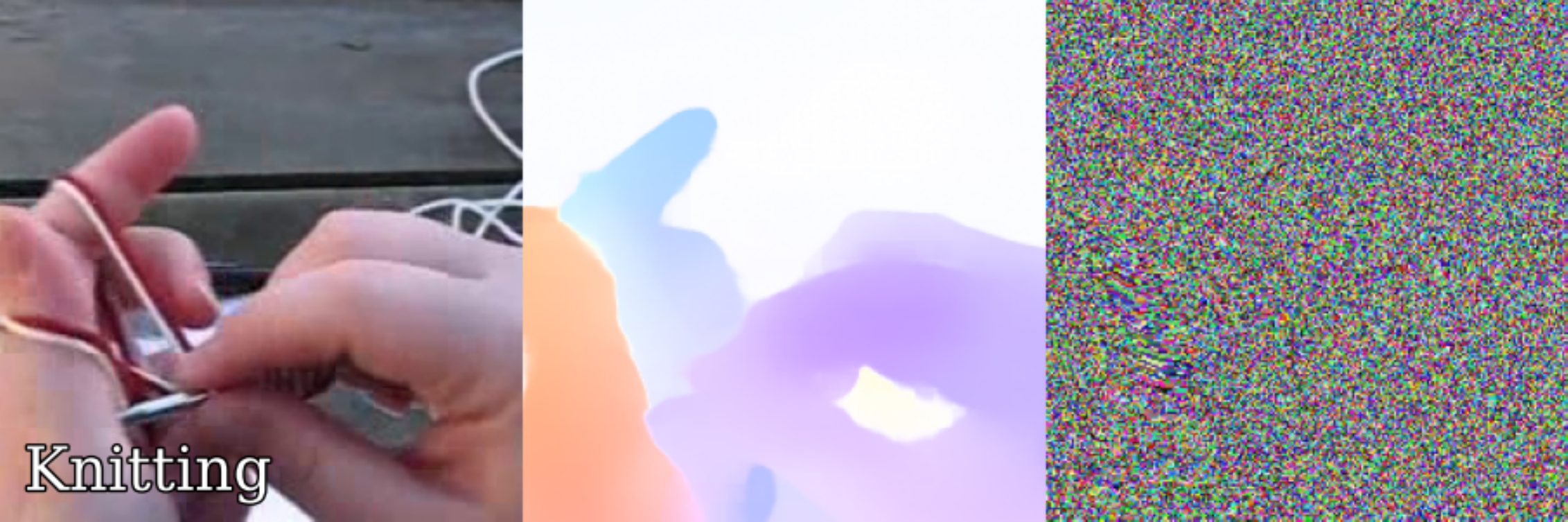}}\\
		\end{tabular}
		\caption{Appearance Free Dataset (AFD) for Action Recognition. 
			Within each set of three images we show: a single RGB frame (left); corresponding optical flow in Middlebury colour coding \cite{baker2011databaseMiddlebury} (middle); corresponding appearance free frame (right). When viewed as video the AFD reveals the motion in the original video even as any single frame provides no relevant discriminative information.
		}
		\label{fig:teaser}
	\end{center}
\end{figure}

\section{Introduction}
\label{sec:intro}
	\let\thefootnote\relax\footnotetext{Code \& Data: \href{https://f-ilic.github.io/AppearanceFreeActionRecognition}{f-ilic.github.io/AppearanceFreeActionRecognition}}
\subsection{Motivation}

\label{sec:motivation}
Action recognition from video has been subject of significant effort in developing and improving both algorithms and datasets \cite{kang2016review,zhu2020comprehensive}.
This interplay between algorithm and dataset advances has led to paradigm shifts in both. Algorithms have evolved from primarily hand-crafted mathematically, physically, and heuristically driven approaches to methods based on deep-learning architectures. Datasets have evolved from relatively small, carefully selected videos to massive, web-crawled collections. As a result, current state-of-the-art algorithms for action recognition achieve impressive levels of performance on challenging datasets. In contrast, the internal representations learned by the architectures remain under explored \cite{hiley2019explainable}. This lack of understanding is unsatisfying both scientifically as well as pragmatically. From a scientific perspective, such detailed analysis is integral to understanding how a system operates and can guide further improvements. From a pragmatic perspective, multiple jurisdictions are beginning to require explainability as a precondition for deployment of artificial intelligence technologies \cite{EU,ON}; technologies lacking such documentation may not see real-world application.

Some evidence suggests that contemporary deep-learning architectures for video understanding can perform well on the task of action recognition with little to no regard for the actual actors \cite{vu2014predicting,he2016human,choi2019can}. These studies suggest that static visual information available in a single frame (\eg~colour, spatial texture, contours, shape) drives performance, rather than dynamic information (\eg,  motion, temporal texture). While some methodologies have been aimed at understanding the representations learned by these architectures \cite{li2018resound,feichtenhofer2020deep,zhao2021interpretable}, none provide an approach to completely disentangle static vs. dynamic information. In response, we have developed an Appearance Free Dataset (AFD) for evaluating action recognition architectures when presented with purely dynamic information; see Fig.~\ref{fig:teaser}. AFD has been rendered by animating spatial noise, historically known as Random Dot Cinematographs RDCs \cite{julesz1971foundations}, using image motion extracted from the UCF101 dataset \cite{soomro2012ucf101} by a state-of-the-art optical flow estimator \cite{teed2020raft}. The resulting videos show no pattern relevant to action recognition in any single frame; however, when viewed as video reveal the underlying motion. We have produced AFD for the entire UCF101 dataset, evaluated a representative sample of contemporary action recognition architectures and used our results to drive development of a novel architecture that enhances performance when confronted with purely dynamic information, as present in AFD.

The ability of action recognition (and other) video  architectures to work in absence of static appearance information is not only an academic exercise. Real-world deployment scenarios may require it, such as in the presence of camouflage. Two examples: Video surveillance for security should be able to cope with nefarious actors who artificially camouflage their activities; video-based wildlife monitoring must be robust to the natural camouflage that many animals posses to hide their presence.

\subsection{Related work}
\label{sec:related}

\paragraph{Datasets} A wide variety of datasets for development and evaluation of automated approaches of video understanding are available  \cite{aafaq2019video}. For action recognition, in particular, there is a large body ranging from a few dozen classes
\cite{kuehne2011hmdb,soomro2012ucf101,d48_web,li2018resound,karpathy2014largesports1m} to massively crawled datasets with classes in the hundreds \cite{karpathy2014largesports1m,carreira2017quo,goyal2017something,gu2018ava}. Some work has specifically focused on curating videos where temporal modeling is particularly important for action recognition \cite{kong2014interactiveBIT,sevilla2021only,goyal2017something,d48_web}; however, they still contain strong cues from single frame static appearance (\eg~colour, shape, texture). Overall, no action recognition dataset completely disentangles single frame static information from multiframe dynamic information.

Camouflaged actors would provide a way to evaluate recognition systems based primarily on dynamic information. While there are camouflage video understanding dataset available, they are of animals in the wild as selected for object segmentation (\eg~\cite{Lamdouar20MocA,bideau2016s}). In contrast, it is unlikely that a non-trivial number of real-world videos of camouflaged actors can be found. In response, our Appearance Free Dataset (AFD) provides synthetic camouflaged action recognition videos as coherently moving patterns of spatial noise, historically known as Random Dot Cinematograms (RDCs) \cite{julesz1971foundations}. These patterns are defined via animation of an initial random spatial pattern, where the pixelwise intensity values of the pattern are randomly selected.
By design, they reveal no information relevant to the motion in any single frame; however, when viewed as video the motion is apparent; examples are shown in the right columns of Fig.~\ref{fig:teaser}. Such videos have a long history in the study of motion processing in both biological \cite{braddick1980low,nishida2018motion} and machine \cite{ullman1979interpretation,zhou2012coherent} vision systems. Interestingly, humans can understand complex human body motion solely from sparse dots marking certain body points, even while merely a random dot pattern is perceived in any single frame \cite{johansson1973visual}.

Synthetic video, both striving for photorealism \cite{mayer2016large,ButlerSintel,dosovitskiy2015flownet,ros2016synthia,richter2017playing}
and more abstracted \cite{sriastava2015,mahmood2019}, is a common tool in contemporary computer vision research that allows for careful control of variables of interest. Our effort adds to this body with its unique contribution of a synthetic camouflage action recognition dataset for probing a system's ability to recognize actions purely based on dynamic information. Notably, while our texture patterns are synthetic, they are animated by the motion of real-world actions \cite{soomro2012ucf101}.

\paragraph{Models and action recognition architectures} A wide range of action recognition architectures have been developed \cite{aafaq2019video}. Most contemporary architectures can be categorized as single stream 3D $(x,y,t)$ convolutional, two-stream and attention based methods. Single stream approaches are motivated by the great success of 2D convolutional architectures on single image tasks \cite{krizhevsky2012imagenet,vgg,iandola2016squeezenet,he2016deep}. In essence, the 3D architectures extend the same style of processing by adding temporal support to their operations \cite{ji20123d,tran2015learning,carreira2017quo,tran2018closer,feichtenhofer2020x3d}. Two-stream architectures have roots in biological vision systems \cite{hubel1959receptive,goodale1992separate}. The idea of having separate pathways for \textbf{static} video content and \textbf{dynamic} information has been variously adopted. A key distinction in these designs is whether the pathway for dynamic information explicitly relies on optical flow estimation \cite{simonyan2014two,feichtenhofer2017spatiotemporal} or  internally computed features that are thought to emulate flow like properties \cite{feichtenhofer2019slowfast}. Attention-based approaches rely on various forms of non-local spatiotemporal data association as manifested in transformer architectures \cite{wang2018non,fan2021multiscale}.  For evaluation of architectures on the new AFD, we select a representative sampling from each of these categories.

\paragraph{Interpretability}

Various efforts have addressed the representational abilities of video understanding architectures ranging from dynamic texture recognition \cite{hadji2018new}, future frame selection \cite{ghodrati2018video} and comparing 3D convolutional vs. LSTM architectures \cite{manttari2020interpreting}. Other work centering on action recognition focused on visualization of learned filters in convolutional architectures \cite{feichtenhofer2020deep,zhao2021interpretable}, or trying to remove scene biases for action recognition \cite{li2018resound,d48_web}. Evidence also suggests that optical flow is useful exactly because it contains invariances related to single frame appearance information \cite{sevilla2018integration}. More closely related to our study is work that categorized various action classes as especially dependent on temporal information; however, single frame static information still remains \cite{sevilla2021only}. Somewhat similarly, an approach tried to tease apart the bias of various architectures and datasets towards static vs. dynamic information by manipulating videos through stylization; however, single frame static information remained a confounding factor \cite{matt2022}.
While insights have been gained, no previous research has been able to completely disentangle the ability of these models to capitalize on single frame static information vs. dynamic information present across multiple frames. 
We concentrate on the ability of these systems to capture dynamic information, with an emphasis on action recognition.

\subsection{Contributions}
\label{sec:contributions}

In light of previous work, we make the following three major contributions.
\begin{enumerate}
	\item We introduce the Appearance Free Dataset (AFD) for video-based action recognition. AFD is a synthetic derivative of UCF101 \cite{soomro2012ucf101} having \textit{no static} appearance cues, however, revealing real-world motion as video.
	\item We evaluate 11 contemporary architectures for action recognition on AFD; \ie~dynamic information alone.
	We conduct a psychophysical study with humans on AFD, and show significanly better performance of humans than networks. These results question the ability of the tested networks to use dynamic information effectively.
	\item We provide a novel improved action recognition architecture with enhanced performance on AFD, while maintaining performance on standard input. 
\end{enumerate}   
\vspace{-5pt}
We make AFD, associated code and our novel architecture publicly available.

\section{Appearance free dataset}
\label{sec:dataset}

Our proposed appearance free dataset, AFD101, is built from the original UCF101 \cite{soomro2012ucf101} by employing a state-of-the-art optical flow estimator \cite{teed2020raft} to generate the corresponding framewise flow that is used to animate spatial noise patterns. The resulting AFD101, consisting of 13,320 videos, depicts realistic motion even while any individual frame is devoid of static appearance information that is relevant for action recognition.
Figure~\ref{fig:teaser} and the project webpage provide illustrative examples.
We use UCF101 from many possible choices \cite{aafaq2019video} as the basis for AFD because it was widely used in action recognition evaluation and is large enough to support training, yet small enough to facilitate numerous experiments.
\begin{figure}[!ht]
	\centering
	\includegraphics[width=0.85\linewidth]{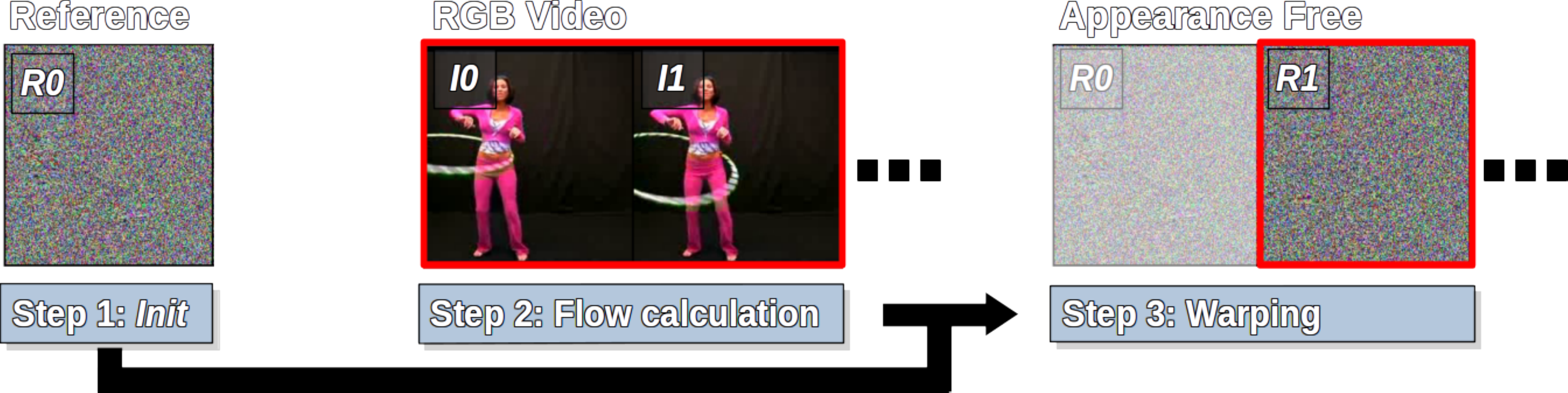}
	\caption{Appearance free videos are created by 1) initializing noise \textit{R0}, 2) calculating optical flow between input frame pairs, and 3) using the optical flow to warp \textit{R0}. Steps 2 and 3 are repeated for each frame-pair in the input video.}
	\label{fig:method}
\end{figure}

\subsection{Dataset generation methodology}
\label{sec:methodologyAFD101}

Generation of AFD from RGB video follows three key steps; see Fig.~\ref{fig:method}.

\begin{enumerate}
	\item Initialization: Generate a single frame of noise, with same spatial dimensions as the frames in the input video. The pixel values are sampled uniformly i.i.d. as three-channel RGB values. Let this frame be denoted as $R_0$.
	\item Flow calculation: Generate interframe optical flow for each temporally adjacent pair of input RGB frames, $I_{t-1}, I_{t}, t\in{[1,T]}$, yielding the flow $F_{t-1,t}$, with $T$ being the number of frames in the input video.
	We use RAFT \cite{teed2020raft}, for the extraction of optical flow.
	\item Warping: We warp the initial noise frame $R_0$ using $F_{t-1,t}$, $t\in{[1,T]}$, to generate the next noise frame, $R_{t}$. 
	The output frames, $R_t$, are an appearance free version of the input RGB video, where at each frame all that is seen is i.i.d. noise, but whose interframe motion corresponds to that of the original video; this output video has the same spatial and temporal dimension as the input video.
\end{enumerate}

\subsection{Implementation details}
\label{sec:implementation}

The details of our proposed methodology are particularly important, as all \textit{static} discriminatory single frame information must be removed; to this end nearest neighbor interpolation during warping is performed. For each warp, newly occluded pixels are overwritten by the new value moved to that location; de-occluded pixels are filled with $R_0$ values to ensure that the spatial extent of the noise is constant. Along the border where the warping leads to undefined values (\eg~pixel at that location moves elsewhere and no new value is warped into that location) are treated as de-occlusions as well. 
In preliminary experiments, we found that other choices of implementation, \eg~bilinear interpolation as opposed to nearest neighbor interpolation, or always estimating flow with respect to $I_0$ always yielded inferior results and unwanted artifacts. In the worst cases these artifacts led to revealing motion boundaries of actors in single frames.

\subsection{Is optical flow appearance free?}

The concept of appearance consists of multiple intertwined qualities, including texture, contour and shape. While optical flow is indeed free of texture it is not free of contour and shape induced by motion of the camera or the actor itself; this fact can be observed qualitatively in Fig.~\ref{fig:teaser}: The flow visualizations alone reveal violin playing and knitting. We quantify this observation as follows. We train a strong single frame recognition architecture, ResNet50 \cite{he2016deep}, on single frames from three datasets: UCF101 (the original), UCF101Flow, comprised of optical flow frames generated from UCF101 by RAFT \cite{teed2020raft}, and AFD101. Top-1 recognition results for UCF101, UCF101Flow and AFD101 are 65.6\%, 29.4\% and 1.1\%, resp. These results show that only AFD101 contains no discriminatory information in a single frame, as only 1.1\% top-1 accuracy is in line with random guessing.

\begin{figure*}[!b]
	\centering
	\includegraphics[width=\linewidth]{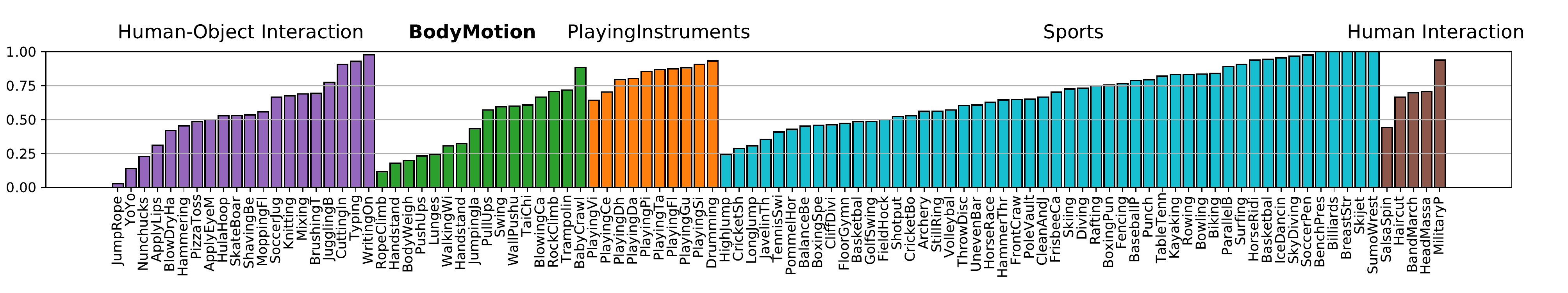}
	\caption{ResNet50 Top-1 Accuracy on single frame UCF101 by \textit{Action Groups}.}
	\label{fig:resnet50UCF101performance}
\end{figure*}

\subsection{AFD5: A reduced subset suitable for small scale exploration}
\label{sec:methodologyAFD5}
To evaluate the performance of state-of-the-art architectures in the context of human performance, a reduced subset of the classes in UCF101 is needed, simply because it is not feasible to conduct an experiment with humans with the original 101 categories. We chose to have 5 classes as it is possible for humans to hold 5 action classes in working memory for a prolonged duration \cite{miller1956magical}. In the following, we refer to the reduced size version of AFD101 as AFD5 and the corresponding reduction of UCF101, \ie~the RGB video, as UCF5.

A defining property of the subset is that it discourages recognition from static information in single frames even within the original RGB input. 
To this end we train a ResNet50 \cite{he2016deep}, used by many state-of-the-art architectures as inspiration for their backbone \cite{aafaq2019video}, on the entire UCF101 dataset in a single frame classification manner, \ie, we randomly sample a frame from UCF101. By design, this procedure isolates the ability of any given class to support recognition on the basis of a single frame. UCF101 is subdivided into five action groups \cite{soomro2012ucf101}: \textit{Human Object Interaction}, \textit{BodyMotion}, \textit{PlayingInstruments}, \textit{Sports} and \textit{Human Human Interaction}; Fig.~\ref{fig:resnet50UCF101performance} has single frame recognition accuracy by group. 

Our selection criterion is to choose videos with similar appearance; therefore, we select a subset of actions within a group. To select the action group we consider the following points. The \textit{Body Motion} group is particularly attractive because it is the only category that has five or more classes with below $25$\% single frame recognition accuracy. The \textit{Human Object Interaction} group is the next closest in terms of having several (albeit fewer that five) categories with such low single frame accuracy. However, its classes tend to be distinguished by featuring a prominent object and we do not want to promote categorization based on single frame object recognition. The \textit{PlayingInstruments} group also features prominent objects and likewise is a poor choice. The remaining groups (\textit{Sports} and \textit{Human Human Interaction}) have generally higher single frame recognition accuracies; so, also are excluded. These considerations lead us to \textit{Body Motion} as the group of interest for present concerns. 

\begin{figure}[tb]
	\begin{tabularx}{\linewidth}{YYY}
		\multicolumn{3}{c}{\includegraphics[width=\linewidth]{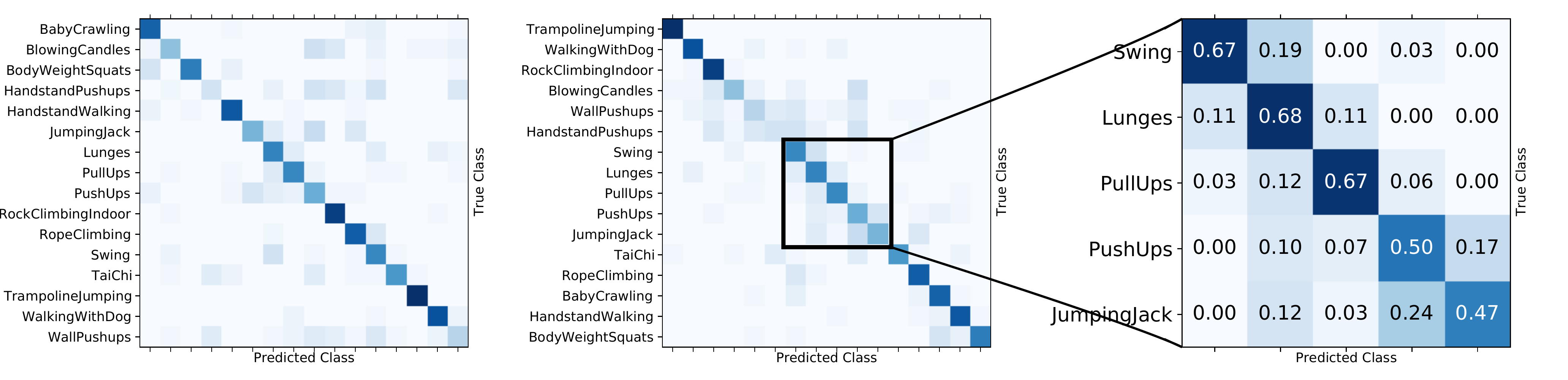}}\\
		a) Alphabetically Sorted& b) Reordered by diagonal & c) Selected Subgroup \\
	\end{tabularx}
	\caption{Singleframe ResNet50: After finetuning on the \textit{BodyMotion} action group of UCF101, we a) inspect the resulting confusion matrix b) reorder it by its diagonal weight so that classes that are often confused among each other get grouped together, and c) select the most prominent subgroup.
	}
	\label{fig:confusionMatrixReordering}
\end{figure}

With the selection of \textit{Body Motion}, we finetune the previously trained ResNet on that particular group. Given that finetuning, we perform confusion matrix reordering \cite{thoma2017analysis} and select the five classes that are most confused among each other in the group. More precisely we find the permutation matrix that has the highest interclass entropy, excluding classes with less than 50\% accuracy; see Fig~\ref{fig:confusionMatrixReordering}. This results in the choice of the following five classes for AFD5: \textit{Swing, Lunges, PushUps, PullUps, JumpingJack}. AFD5 consists in total of $583$ videos with approximately $116$ videos per class.

\section{Psychophysical study: Human performance on AFD5}
\label{sec:humans}

To set a baseline for Appearance Free Data we perform a psychophysical study.
The ability of humans to recognize actions  from dynamic information alone, \ie~without any single frame static information, has been documented previously \cite{johansson1973visual,dittrich1993action,troje2002decomposing}. These studies are conducted with sparse dots, typically at major body joints, on otherwise blank displays. Our study appears to be unique in its use of dense noise patterns~\ie~dense random dots.

\subsection{Experiment design}
\label{sec:humanDesign}

\begin{figure*}[tb]
	\includegraphics[trim=0 88 0 0,clip,width=\linewidth]{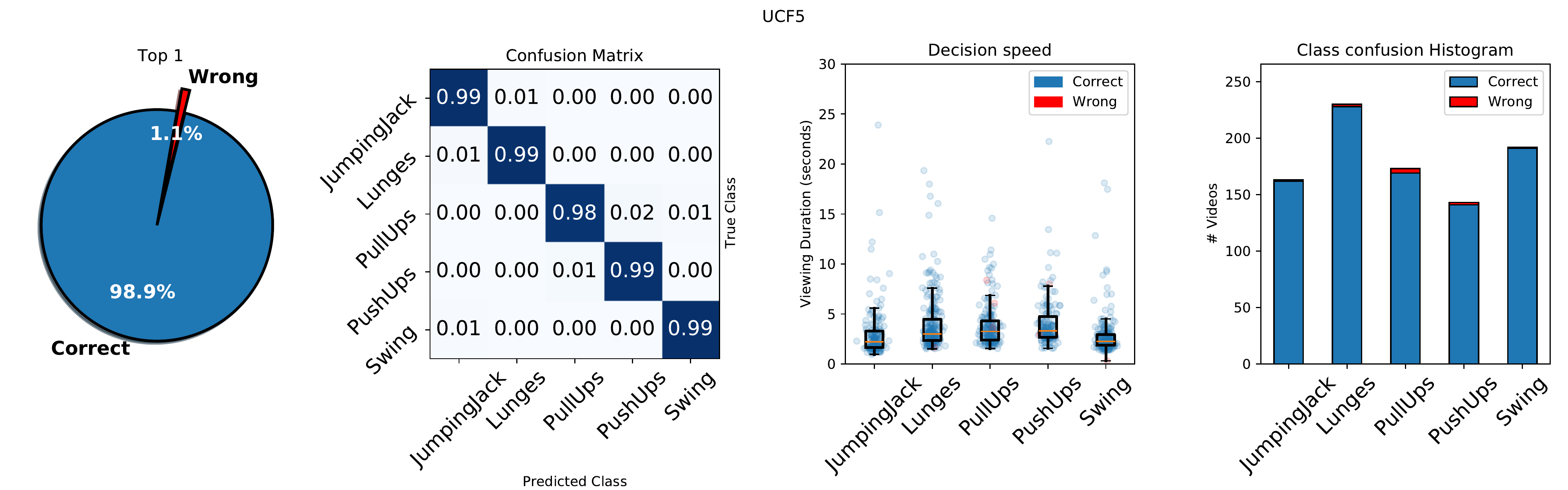}
	\includegraphics[width=\linewidth]{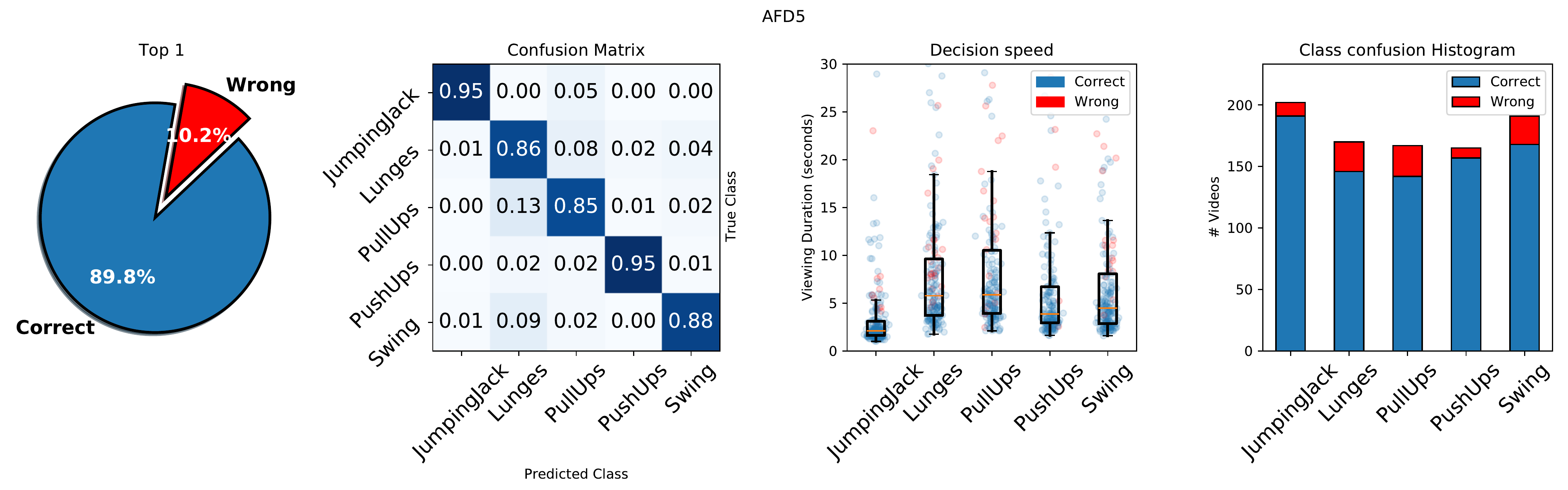}
	\begin{tabularx}{\linewidth}{YYYY}
		Average Top1 Accuracy& Overall Confusion Matrix & Response Times per Class & Class Confusion Histogram\\
	\end{tabularx}
	\caption{Results of psychopysical study. Top row: Performance on standard RGB video, UCF5. Bottom row: Performance on appearance free data, AFD5. 
	}
	\label{fig:userstudy}
\end{figure*}

A session in the experiment consisted of the following seven phases. (i) The participant is presented with a brief slide show explaining the task. The participant is told that they will see a series of videos as well as a menu on the same display from which to indicate their perception of a human action that is depicted. (ii) Training is provided on UCF5, \ie, RGB video. During this phase, four training videos from each class, totaling 20 videos, are presented with correct responses indicated in the menu. The videos are shown in randomized order. (iii) Testing is performed on 90 UCF5 test videos. Each video repeats until the participant makes their selection on the menu. (iv) A rest period of five minutes is provided. (v) Training is provided on AFD5, \ie~appearance free video. During this phase, four training videos from each class are presented with correct responses indicated in the menu. The videos also are shown in randomized order. (vi) Testing is performed on 90 AFD5 test videos. Each video repeats until the participant makes their selection on the menu. (vii) Participants are given a questionnaire to complete, with questions including visual impairments (all had normal or corrected vision), familiarity with computer vision, action recognition, as well as their impression of task difficulty. During both test phases, action category choices and response times were recorded. 
Ten human participants were recruited as volunteers to participate in the experiment. All experiments were conducted in the same environment, \ie~computer/display, lighting and seating distance from the display (0.5 meters). Participants were allowed to view freely from their seat and take as much time in making their responses as they like. All participants completed their session within approximately 30 minutes, including the questionnaire. Participants were informed and consented to the data gathered in written and signed manner, and were provided with sweets and thanks at task completion. The sample size of ten participants is in our view sufficient to show that various people with different backgrounds are \textit{able} to solve AFD data without much trouble and to establish a baseline for comparison to action recognition architectures.

\subsection{Human results}
\label{sec:humanResults}
Results of the psychophysical study are shown in Fig.~\ref{fig:userstudy}. On average, our participants perform with an accuracy of 89.8\% on AFD5 and 98.9\% on UCF5. Participants report that mistakes on UCF5, the RGB videos, resulted from accidental clicks when making selections. Our analysis of AFD5 samples that were most frequently misclassified by the participants reveals that those videos typically have sudden and high speed camera movement, \eg~as induced by quick panning and zooming, which in turn creates very large optical flow displacement.
The plots of response time show that participants take longer to make their choice on AFD5, with the exception of \textit{JumpingJack} which remains largely unaffected.  The extended time taken for AFD5 suggests that appearance free recognition requires more effort and repeated playback of the video is necessary, even though ultimate performance is high. 

Overall, the psychophysical  results document the strong ability of humans to recognize actions on the basis of AFD, as well as standard RGB videos.

\section{Computational study: Model performances on AFD}
\label{sec:machine}
We evaluate 11 state-of-the-art action recognition architectures on UCF101 and our new AFD101. We select representative examples from the currently most widely used categories of action recognition architectures, \textit{cf}. Sec.~\ref{sec:related}: single stream 3D convolutional \cite{feichtenhofer2020x3d,carreira2017quo,tran2018closer}, two-stream convolutional \cite{simonyan2014two,feichtenhofer2019slowfast} and attention-based \cite{fan2021multiscale,wang2018non}. We also evaluate on a standard 2D convolutional architecture (ResNet50 \cite{he2016deep}) to verify that our AFD does not support classification based on single frame static information.
Notably, our results from the psychophysical study allow us to compare the performance of the network architectures to that of humans on both UCF5 and AFD5.

\subsection{Training procedure}
To guarantee a fair comparison all reported results are obtained by training and testing with our procedure, described below.
Furthermore, since we want to investigate architectural benefits on Appearance Free Data, all results (except for one architecture, see below) are obtained without pretraining - concurrently often done on Kinetics400 \cite{carreira2017quo} - as there is no AFD for Kinetics. Our goal is not to show how to archieve best performance on UCF101; it is to show the difference in performance of each network when presented with \textit{only} dynamic information. We also ran preliminary experiments with pretraining on Kinetics and finetuning on AFD, but saw no significant improvement during AFD testing compared to no pretraining. Nevertheless, since all the evaluated architectures, save one, employ a ResNet-like backbone we use ImageNet pretrained versions of those to initialize our training. The exception is MViT, which needed to be initialized with Kintecs400 weights, as its performance was not increasing from baseline with the aforementioned scheme. However the Kinetics weights were only used as an intializer and we allowed for retraining of the entire backbone to make the results comparing UCF and AFD as comparable as possible. We acknowledge that our training strategy does not allow for the architectures to show state-of-the-art performance on RGB input in our experiments; however, our approach is necessary to allow for fair comparison of performance on RGB vs. AFD and measure the respective performance between the modalities.

\subsection{Training details}
We train with Adam \cite{adam} using an early stopping scheme, and a base learning rate of $3e^{-4}$.
The batch size is chosen as large as possible on two GPUs with 24GB memory each. 
Data augmentation, a staple of modern deep learning, is kept the same for all evaluated models; uniform temporal sampling with jitter, random start frame selection, and contrast, hue, and brightness jitter are applied. Random horizontal flipping of entire videos is used, as actions in UCF101 do not depend on handedness. A quite aggressive data augmentation scheme that changes over the course of every 20 epochs to a less aggressive one, and capping at 80 epochs is used. We find that this aggressive data augmentation technique is needed to achieve reasonable results when no pretraining is used. Spatial center crops of 224$\times$224 are used as the final input to all networks with the exception of X3D XS and S, which use a spatial resolution of 160$\times$160.
The reported results are the averages of the 3 standard splits used on UCF101, now applied to both the original RGB as well as AFD.
The results on UCF5 and AFD5 were obtained by finetuning the networks previously trained on their 101 class counterparts and using the same splits.

\subsection{Architecture results}
\label{sec:architectureResults}

Table~\ref{tab:results} shows top-1 accuracy results for all evaluated architectures on UCF101, AFD101, UCF5 and AFD5. Note that the results for the single frame ResNet on UCF101 essentially are the same as those shown in Fig.~\ref{fig:resnet50UCF101performance}. 
The results in Table~\ref{tab:results} also validate our claim of having removed all \textit{static}, single frame information relevant to recognition, as the performance is equal to random choice.

For most of the action recognition architectures, it is seen that obtaining $\approx 70$-$80$\% top-1 accuracy is possible. This fact makes the sizable drop of performance on AFD data, the $\Delta$ columns in Table \ref{tab:results}, especially striking. The average performance drops by approximately 30\% almost evenly across all architectures - with two notable exceptions: The Fast stream of SlowFast and I3D OF a common I3D architecture with an optical flow estimator before the network input. These two architectures are noteworthy because their ability to maintain performance across regular and appearance free video apparently stems from their design to prioritize temporal information over spatial; although, this focus on temporal information leads them to be less competitive on the standard RGB input in comparison to the other evaluated architectures. The comparison to human performance shows X3D variations to be the best architectures, competitive with human performance on UCF5; \textit{however}, the best architectures on AFD5 (X3D M and SlowFast) are \textit{considerably} below human top-1 accuracy (18\% below). These results show that there is room for improvement by enhancing the ability of current architectures to exploit dynamic information for action recognition. It also suggests that optical flow is \textit{not} what is learned by these networks, when no explicit representation of flow is enforced.
Importantly, TwoStream \cite{simonyan2014two} and I3D\cite{carreira2017quo} OF were evaluated  with RAFT optical to allow a for a fair comparison with each other and with our new algorithm, introduced in the next section.
The summary plots below Table~\ref{tab:results} show learnable architecture parameter counts and ordering by AFD accuracy. 
These plots further emphasize the fact that explicit representation of dynamic information (\eg~as in I3D OF and to some extent Fast) is best at closing the gap between performance on RGB and AFD. X3D M is among the top performing architectures on RGB; further, among those top RGB performers, it shows the smallest $\Delta$. Also, it is has a relatively small number of learnable parameters, with only Fast notably smaller. These observations motivate the novel architecture that we present in the next section. It improves performance on AFD, while also slightly improving performance on RGB. This architecture is given as E2S-X3D in Table~\ref{tab:results}. Moreover, it performs on par with humans on UCF5 and AFD5.

\begin{table}[!htb]
	\small
	\begin{center}
		\begin{tabularx}{\linewidth}{lY|YYY|YYY}
			\toprule
			&  $f \times r$ & UCF101  & AFD101 & $\Delta$ & UCF5  & AFD5 & $\Delta$\\ 
			\midrule
			\textbf{Single Image Input}& & &  & &\\
			ResNet50 \cite{he2016deep}  &$1\times1$ & 65.6    & 1.1 &64.5& 74.2  & 20.1&54.1\\
			\midrule
			\textbf{Video Input}& & &&  & &&\\
			I3D \cite{carreira2017quo} &$8\times8$                             & 67.8   &  43.3& 24.5& 95.6  & 55.1& 40.5 \\
			TwoStream \cite{simonyan2014two} & \Arch{1 \times 1}{10 \times 1}  & 76.1   &  29.3  & 46.8 & 78.1 & 62.3 & 15.8 \\
			C2D \cite{wang2018non} &$8\times8$                                 & 77.4   & 42.6&34.8 & 84.5   & 69.0&15.5\\
			R2+1D \cite{tran2018closer}           &$16 \times 4$               & 68.2   &     28.9  & 39.3  & 80.6 &  67.9 & 12.7\\
			Slow\cite{feichtenhofer2019slowfast} &$8\times8$                   & 73.3  & 37.6& 35.7& 89.9   & 40.6&49.3\\ 
			Fast \cite{feichtenhofer2019slowfast} &$32 \times 2$                & 50.5   & 45.9  &4.6&    71.4 & 65.7&5.7\\ 
			SlowFast \cite{feichtenhofer2019slowfast} $\binom{\text{Slow}}{\text{Fast}}$ &$\binom{8\times8}{32\times2}$ & 82.9  & 55.4 &27.5& 91.0  & 70.2&20.8\\
			I3D\cite{carreira2017quo} OF  &$8\times8$                             & 54.5   & 44.4 &10.1 &  74.4     &  59.7 & 14.7\\
			X3D XS \cite{feichtenhofer2020x3d} & $4\times12$                   & 79.8     & 43.2& 36.6& 95.2  & 57.0&38.2\\
			X3D S \cite{feichtenhofer2020x3d} & $13\times6$                    & 80.8   & 56.9 &23.9& 97.3  & 69.8&27.5\\
			X3D M \cite{feichtenhofer2020x3d} & $16\times5$                    & 81.5   & 58.2 &23.3& 98.8 & 71.5&27.3\\
			MViT \cite{fan2021multiscale} & $16 \times 4$ & 82.9   &    39.8  & 43.1&  95.1  & 22.2&72.9\\			
			E2S-X3D \textbf{Ours} \Arch{\text{M}}{\text{S}}&\Arch{16\times5}{13\times6}  & 85.7   & 73.9 &11.8& 99.4 & 90.8&8.6\\
			\midrule
			Human Average  &&& & &   98.9  & 89.8 &9.1\\
			\bottomrule
		\end{tabularx}
		\includegraphics[width=\linewidth]{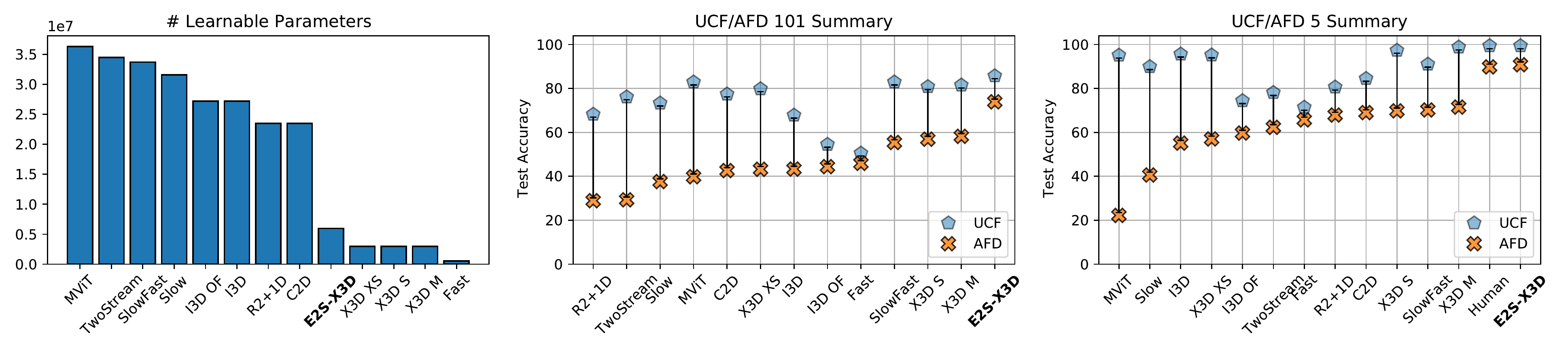}
		\caption{Top-1 accuracy for the evaluated action recognition architectures. The second column indicates the number of frames and the temporal sampling rate, $f \times r$, of each network. For two-stream approaches the notation \Arch{\text{Appearance}}{\text{Motion}}\Arch{f \times r}{f \times r} is used.
			A drop in performance across all networks on AFD data is observed. Plots below the table show parameter counts as well as summarize the accuracy findings, sorted by AFD performance. Absolute performance on UCF is of secondary importance; it merely gives a point of comparison with respect to AFD. The $\Delta$ columns shows the difference between UCF and AFD performance, which highlights the relative \textit{dynamic} recognition capability of architectures.}
		\label{tab:results}
	\end{center}
\end{table}

\section{Two-stream strikes back}
\label{sec:twoStream}

The results of our evaluation of architectures in Sec.~\ref{sec:machine} show the importance of explicit representation of dynamic information for strong performance on AFD. This result motivates us to revive the two-stream appearance plus optical flow design \cite{simonyan2014two}. While state-of-the art when introduced, compared to more recent action recognition architectures, \eg~those evaluated in Sec.~\ref{sec:machine}, it is no longer competitive \cite{zhu2020comprehensive}; so, we incorporate two-streams into the top performing architecture on RGB input with the smallest drop on AFD, \ie~X3D.

\subsection{E2S-X3D: Design of a novel action recognition architecture}	

\begin{figure}[h]
	\centering
	\includegraphics[width=0.85\linewidth]{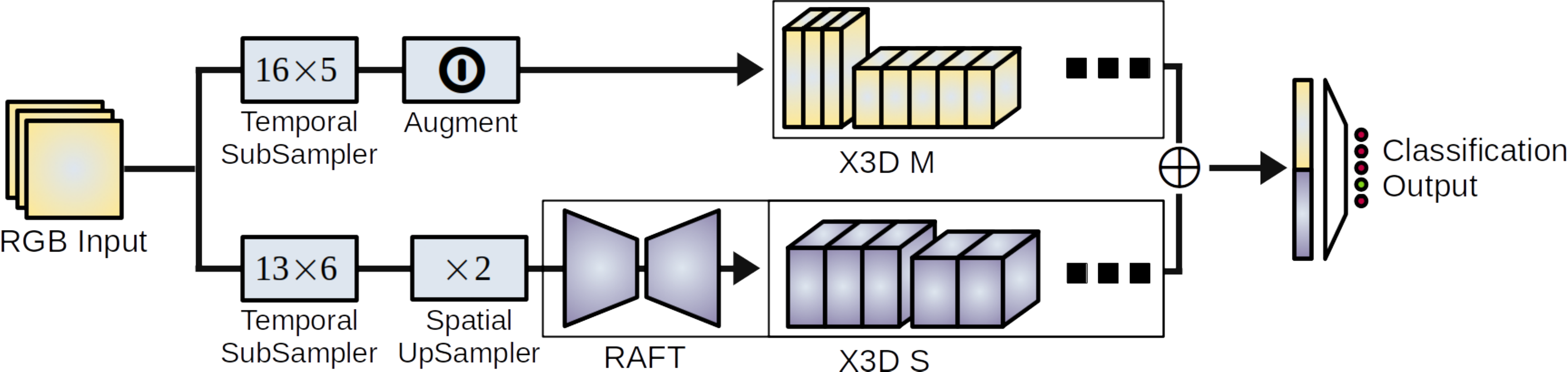}\\
	\resizebox{\linewidth}{!}{
		\begin{tabular}{cc|c|c|c|c|c|c}
			\toprule
			Model & input      & conv$_1$      & res$_2$    & res$_3$     & res$_4$     & res$_5$   & conv$_5$   \\
			\midrule
			\textbf{M} & $16\times224\times224 $ & $16\times112\times112 $ & $16\times56\times56 $ & $16\times28\times28 $ & $16\times14\times14 $ & $16\times7\times7 $ & $16\times7\times7 $  \\
			\textbf{S} & $13\times160\times160 $ & $13\times80\times80   $ & $13\times40\times40 $ & $13\times20\times20 $ & $13\times10\times10 $ & $13\times5\times5 $ & $13\times5\times5 $ \\
			\bottomrule
		\end{tabular}
	}
	\caption{Our novel ES2-X3D operates with two parallel streams for processing of RGB (top stream) and optical flow (bottom stream). Late fusion via concatenation is followed by a fully connected and Softmax layer. The table shows the output sizes of  each employed X3D architecture in the format T$\times$H$\times$W.
	}
	\label{fig:e2sx3ds}
\end{figure}

The design of our novel architecture, ES2-X3D, is shown in Fig.~\ref{fig:e2sx3ds}. Initially, optical flow fields and RGB images are processed in separate 3D convolutional streams. Both streams are versions of the X3D architecture \cite{feichtenhofer2020x3d}, where we use the S variant for optical flow and the M variant for RGB. These two architectures differ in their spatiotemporal resolution of the input and as a consequence the activation maps. X3D M uses larger spatial and temporal extents, whereas S has smaller spatial and temporal supports. Since the input flow field already represents the dynamic information with detail, the additional spatiotemporal extent is not needed; in contrast, the RGB stream without the optical flow benefits from larger spatial and temporal support, see Fig.~\ref{fig:e2sx3ds} (bottom). We validate these choices below via ablation. The dimensionality of the layers is shown in Fig.~\ref{fig:e2sx3ds}. Following the separate parallel processing, late fusion is performed. We fuse the outputs via simple concatenation of the outputs of the two streams. While more sophisticated fusion schemes might be considered, \eg~\cite{feichtenhofer2017spatiotemporal}, we leave that for future work. Finally, a fully connected layer followed by a Softmax layer yields the classification output, which is used with a cross-entropy loss to train the network.
For optical flow input, we use a state-of-the-art optical flow estimator, RAFT \cite{teed2020raft} pretrained on Sintel \cite{ButlerSintel}. We do not further train the flow extractor and only use it in its inference mode. To capture fine grained motion and to adjust for the different spatial resolution RAFT was trained with, we add an up-sampling layer to the RGB images prior to RAFT. Data augmentation, as described in Sec.~\ref{sec:machine}, is only used in the appearance stream, as indicated in Fig.~\ref{fig:e2sx3ds}.
The training, validation and testing of the network follows the same procedure as the other networks; described in Sec.~\ref{sec:machine}.

\begin{table}[!htb]
	\small
	\begin{minipage}{.5\linewidth}
		\centering
		\begin{tabularx}{.9\textwidth}{lcYYY}
			\multicolumn{5}{c}{\textbf{UCF5}}\\
			\toprule
			&& \multicolumn{3}{c}{\textbf{Input}}\\
			\multicolumn{2}{c}{Architecture} & $\binom{\text{{RGB}}}{\text{-}}$  & $\binom{\text{{-}}}{\text{{RAFT}}}$ & $\binom{\text{{RGB}}}{\text{{RAFT}}}$ \\
			\midrule
			XS &$4\times12$ & 95.2 & 82.1 & ---\\
			S &$13\times6$ & 97.3 & \textbf{89.5} & ---\\
			M &$16\times5$ & \textbf{98.8} & 86.8 & ---\\
			\midrule 			
			\Arch{\text{M}}{\text{S}}&\Arch{16\times5}{13\times6} & --- & --- & \textbf{99.4} \\
			\bottomrule
		\end{tabularx}
	\end{minipage}%
	\begin{minipage}{.5\linewidth}
		\centering
		\begin{tabularx}{.9\textwidth}{lcYYY}
			\multicolumn{5}{c}{\textbf{AFD5}}\\
			\toprule
			&& \multicolumn{3}{c}{\textbf{Input}}\\
			\multicolumn{2}{c}{Architecture} & $\binom{\text{{RGB}}}{\text{-}}$  & $\binom{\text{{-}}}{\text{{RAFT}}}$ & $\binom{\text{{RGB}}}{\text{{RAFT}}}$ \\
			\midrule
			XS &$4\times12$ & 57.0 & 77.3 & ---\\
			S &$13\times6$ & 69.8 & \textbf{80.3} & ---\\
			M &$16\times5$ & \textbf{71.5} & 78.9 & ---\\
			\midrule 			
			\Arch{\text{M}}{\text{S}}&\Arch{16\times5}{13\times6} & --- & --- & \textbf{90.8} \\
			\bottomrule
		\end{tabularx}
	\end{minipage}	
	\vspace{5pt}
	\caption{Performance of X3D architecture configurations, XS, S and M for UCF5 and AFD5. Ablation is performed for both optical flow and RGB inputs. 
	}
	\label{tab:ablation}
\end{table}

\subsection{E2S-X3D: Empirical evaluation}
E2S-X3D outperforms all competing architectures on AFD101, with the closest competitor (singlestream X3D) trailing by 15.7\%; see Table~\ref{tab:results}. It also improves performance on UCF101, albeit slightly; 2.8\% points gained on MViT (MViT has around 6$\times$ more parameters). 
To rule out an ensembling effect, we compared our network in its default configuration (RGB in one stream, optical flow in the other) with an alternative where it inputs RGB to both streams.  We find a performance drop of 10.6\% on UCF101 and 30.7\% on AFD101, validating our hypothesis that explicit optical flow is crucial.
Moreover, it is on par with human performance on both AFD5 and UCF5. These results document the advantage of explicit modeling of motion for action recognition.
Table~\ref{tab:ablation} shows results of ablations across various X3D configurations for both AFD5 and UCF5. It shows that for both input modalities, best performance is achieved when the S configuration is used for optical flow and the M configuration for the RGB stream. This set of experiments empirically validates our final ES2-X3D design.

\clearpage
\section{Conclusions}
We introduce an extension to a widely used action recognition dataset that disentangles static and dynamic video components. In particular, no single frame contains any \textit{static} discriminatory information in our apperance free dataset (AFD); the action is only encoded in the temporal dimension. We show, by means of a psychophysical study with humans, that solving AFD is possible with a high degree of certainty ($\sim90$\% Top1 Accuracy). This result is especially interesting as 11 current state-of-the-art action recognition architectures show much weaker performance as well as a steep drop in performance when comparing standard RGB to AFD input. These results lend to the interpretability of the evaluated architectures, by documenting their ability to exploit dynamic information.
In particular, this shows that optical flow, or a similarly descriptive representation, is not an emergent property of \textit{any} of the tested network architectures.

We propose a novel architecture incorporating the explicit computation of optical flow and use insights from recent action recognition research. This explict form of modeling \textit{dynamics} allows our approach to outperform all competing methods, and compete with human level performance. Given the strong performance of our architecture in our evaluation, future work could consider training and investigation of performance on larger datasets \cite{carreira2017quo,goyal2017something} as well as application to other other tasks where non-synthetic data is available and dynamic information plays a crucial role, \eg~camouflaged animal detection \cite{Lamdouar20MocA}. Alternative fusion strategies are also a potential for improvement of our present architecture.

\clearpage
\bibliographystyle{IEEEtran}  
\bibliography{IEEEabrv,references}

\begin{thebibliography}{10}
\providecommand{\url}[1]{#1}
\csname url@samestyle\endcsname
\providecommand{\newblock}{\relax}
\providecommand{\bibinfo}[2]{#2}
\providecommand{\BIBentrySTDinterwordspacing}{\spaceskip=0pt\relax}
\providecommand{\BIBentryALTinterwordstretchfactor}{4}
\providecommand{\BIBentryALTinterwordspacing}{\spaceskip=\fontdimen2\font plus
\BIBentryALTinterwordstretchfactor\fontdimen3\font minus
  \fontdimen4\font\relax}
\providecommand{\BIBforeignlanguage}[2]{{%
\expandafter\ifx\csname l@#1\endcsname\relax
\typeout{** WARNING: IEEEtran.bst: No hyphenation pattern has been}%
\typeout{** loaded for the language `#1'. Using the pattern for}%
\typeout{** the default language instead.}%
\else
\language=\csname l@#1\endcsname
\fi
#2}}
\providecommand{\BIBdecl}{\relax}
\BIBdecl

\bibitem{baker2011databaseMiddlebury}
S.~Baker, D.~Scharstein, J.~Lewis, S.~Roth, M.~J. Black, and R.~Szeliski, ``A
  database and evaluation methodology for optical flow,'' \emph{International
  Journal of Computer Vision}, vol.~92, no.~1, pp. 1--31, 2011.

\bibitem{kang2016review}
S.~M. Kang and R.~P. Wildes, ``Review of action recognition and detection
  methods,'' \emph{arXiv preprint arXiv:1610.06906}, 2016.

\bibitem{zhu2020comprehensive}
Y.~Zhu, X.~Li, C.~Liu, M.~Zolfaghari, Y.~Xiong, C.~Wu, Z.~Zhang, J.~Tighe,
  R.~Manmatha, and M.~Li, ``A comprehensive study of deep video action
  recognition,'' \emph{arXiv preprint arXiv:2012.06567}, 2020.

\bibitem{hiley2019explainable}
L.~Hiley, A.~Preece, and Y.~Hicks, ``Explainable deep learning for video
  recognition tasks: A framework \& recommendations,'' \emph{arXiv preprint
  arXiv:1909.05667}, 2019.

\bibitem{EU}
``Laying down harmonised rules on artificial intelligence (artificial
  intelligence act) and amending certain union legislative acts,'' European
  Commision, 2021.

\bibitem{ON}
``Regulating {AI}: Critical issues and choices,'' Law Council of Ontario, 2021.

\bibitem{vu2014predicting}
T.-H. Vu, C.~Olsson, I.~Laptev, A.~Oliva, and J.~Sivic, ``Predicting {A}ctions
  from {S}tatic scenes,'' in \emph{Proceedings of the European Conference on
  Computer Vision}, 2014.

\bibitem{he2016human}
Y.~He, S.~Shirakabe, Y.~Satoh, and H.~Kataoka, ``Human action recognition
  without human,'' in \emph{Proceedings of the European Conference on Computer
  Vision}, 2016.

\bibitem{choi2019can}
J.~Choi, C.~Gao, C.~E.~J. Messou, and J.-B. Huang, ``Why can't {I} dance in the
  mall? {L}earning to mitigate scene bias in action recognition,'' in
  \emph{Proceedings of the Conference on Advances in Neural Information
  Processing Systems}, 2019.

\bibitem{li2018resound}
Y.~Li, Y.~Li, and N.~Vasconcelos, ``Resound: Towards action recognition without
  representation bias,'' in \emph{Proceedings of the European Conference on
  Computer Vision}, 2018.

\bibitem{feichtenhofer2020deep}
C.~Feichtenhofer, A.~Pinz, R.~P. Wildes, and A.~Zisserman, ``Deep insights into
  convolutional networks for video recognition,'' \emph{International Journal
  of Computer Vision}, vol. 128, no.~2, pp. 420--437, 2020.

\bibitem{zhao2021interpretable}
H.~Zhao and R.~P. Wildes, ``Interpretable deep feature propagation for early
  action recognition,'' \emph{arXiv preprint arXiv:2107.05122}, 2021.

\bibitem{julesz1971foundations}
B.~Julesz, \emph{Foundations of Cyclopean Perception.}\hskip 1em plus 0.5em
  minus 0.4em\relax U. Chicago Press, 1971.

\bibitem{soomro2012ucf101}
K.~Soomro, A.~R. Zamir, and M.~Shah, ``{UCF101}: A dataset of 101 human actions
  classes from videos in the wild,'' \emph{arXiv preprint arXiv:1212.0402},
  2012.

\bibitem{teed2020raft}
Z.~Teed and J.~Deng, ``{RAFT}: Recurrent all-pairs field transforms for optical
  flow,'' in \emph{Proceedings of the European Conference on Computer Vision},
  2020.

\bibitem{aafaq2019video}
N.~Aafaq, A.~Mian, W.~Liu, S.~Z. Gilani, and M.~Shah, ``Video description: A
  survey of methods, datasets, and evaluation metrics,'' \emph{ACM Computing
  Surveys}, vol.~52, no.~6, pp. 1--37, 2019.

\bibitem{kuehne2011hmdb}
H.~Kuehne, H.~Jhuang, E.~Garrote, T.~Poggio, and T.~Serre, ``{HMDB}: {A} large
  video database for human motion recognition,'' in \emph{Proceedings of the
  International Conference on Computer Vision}, 2011.

\bibitem{d48_web}
Y.~Li, Y.~Li, and N.~Vasconcelos, ``Diving48 dataset,''
  \url{http://www.svcl.ucsd.edu/projects/resound/dataset.html}.

\bibitem{karpathy2014largesports1m}
A.~Karpathy, G.~Toderici, S.~Shetty, T.~Leung, R.~Sukthankar, and L.~Fei-Fei,
  ``Large-scale video classification with convolutional neural networks,'' in
  \emph{Proceedings of the Conference on Computer Vision and Pattern
  Recognition}, 2014.

\bibitem{carreira2017quo}
J.~Carreira and A.~Zisserman, ``Quo vadis, action recognition? {A} new model
  and the kinetics dataset,'' in \emph{Proceedings of the Conference on
  Computer Vision and Pattern Recognition}, 2017.

\bibitem{goyal2017something}
R.~Goyal, S.~Ebrahimi~Kahou, V.~Michalski, J.~Materzynska, S.~Westphal, H.~Kim,
  V.~Haenel, I.~Fruend, P.~Yianilos, M.~Mueller-Freitag, F.~Hoppe, C.~Thurau,
  I.~Bax, and R.~Memisevic, ``The ``something something" video database for
  learning and evaluating visual common sense,'' in \emph{Proceedings of the
  International Conference on Computer Vision}, 2017.

\bibitem{gu2018ava}
C.~Gu, C.~Sun, D.~A. Ross, C.~Vondrick, C.~Pantofaru, Y.~Li,
  S.~Vijayanarasimhan, G.~Toderici, S.~Ricco, R.~Sukthankar \emph{et~al.},
  ``{AVA}: A video dataset of spatio-temporally localized atomic visual
  actions,'' in \emph{Proceedings of the Conference on Computer Vision and
  Pattern Recognition}, 2018.

\bibitem{kong2014interactiveBIT}
Y.~Kong, Y.~Jia, and Y.~Fu, ``Interactive phrases: Semantic descriptions for
  human interaction recognition,'' \emph{IEEE Transactions on Pattern Analysis
  and Machine Intelligence}, vol.~36, no.~9, pp. 1775--1788, 2014.

\bibitem{sevilla2021only}
L.~Sevilla-Lara, S.~Zha, Z.~Yan, V.~Goswami, M.~Feiszli, and L.~Torresani,
  ``Only time can tell: Discovering temporal data for temporal modeling,'' in
  \emph{Proceedings of the Winter Conference on Applications of Computer
  Vision}, 2021.

\bibitem{Lamdouar20MocA}
H.~Lamdouar, C.~Yang, W.~Xie, and A.~Zisserman, ``Betrayed by motion:
  Camouflaged object discovery via motion segmentation,'' in \emph{Proceedings
  of the Asian Conference on Computer Vision}, 2020.

\bibitem{bideau2016s}
P.~Bideau and E.~Learned-Miller, ``It’s moving! {A} probabilistic model for
  causal motion segmentation in moving camera videos,'' in \emph{Proceedings of
  the European Conference on Computer Vision}, 2016.

\bibitem{braddick1980low}
O.~J. Braddick, ``Low-level and high-level processes in apparent motion,''
  \emph{Philosophical Transactions of the Royal Society of London. B,
  Biological Sciences}, vol. 290, no. 1038, pp. 137--151, 1980.

\bibitem{nishida2018motion}
S.~Nishida, T.~Kawabe, M.~Sawayama, and T.~Fukiage, ``Motion perception: From
  detection to interpretation,'' \emph{Annual review of Vision Science},
  vol.~4, pp. 501--523, 2018.

\bibitem{ullman1979interpretation}
S.~Ullman, \emph{The Interpretation of Visual Motion}.\hskip 1em plus 0.5em
  minus 0.4em\relax MIT Press, 1979.

\bibitem{zhou2012coherent}
B.~Zhou, X.~Tang, and X.~Wang, ``Coherent filtering: Detecting coherent motions
  from crowd clutters,'' in \emph{Proceedings of the European Conference on
  Computer Vision}, 2012.

\bibitem{johansson1973visual}
G.~Johansson, ``Visual perception of biological motion and a model for its
  analysis,'' \emph{Perception \& Psychophysics}, vol.~14, no.~2, pp. 201--211,
  1973.

\bibitem{mayer2016large}
N.~Mayer, E.~Ilg, P.~Hausser, P.~Fischer, D.~Cremers, A.~Dosovitskiy, and
  T.~Brox, ``A large dataset to train convolutional networks for disparity,
  optical flow, and scene flow estimation,'' in \emph{Proceedings of the
  Conference on Computer Vision and Pattern Recognition}, 2016.

\bibitem{ButlerSintel}
D.~J. Butler, J.~Wulff, G.~B. Stanley, and M.~J. Black, ``A naturalistic open
  source movie for optical flow evaluation,'' in \emph{Proceedings of the
  European Conference on Computer Vision}, 2012.

\bibitem{dosovitskiy2015flownet}
A.~Dosovitskiy, P.~Fischer, E.~Ilg, P.~Hausser, C.~Hazirbas, V.~Golkov, P.~Van
  Der~Smagt, D.~Cremers, and T.~Brox, ``Flow{N}et: Learning optical flow with
  convolutional networks,'' in \emph{Proceedings of the International
  Conference on Computer Vision}, 2015.

\bibitem{ros2016synthia}
G.~Ros, L.~Sellart, J.~Materzynska, D.~Vazquez, and A.~M. Lopez, ``The
  {SYNTHIA} dataset: A large collection of synthetic images for semantic
  segmentation of urban scenes,'' in \emph{Proceedings of the Conference on
  Computer Vision and Pattern Recognition}, 2016.

\bibitem{richter2017playing}
S.~R. Richter, Z.~Hayder, and V.~Koltun, ``Playing for benchmarks,'' in
  \emph{Proceedings of the International Conference on Computer Vision}, 2017.

\bibitem{sriastava2015}
N.~Sriastava, E.~Manisomov, and R.~Salakhutdinov, ``Unsupervised learning of
  video representations using {LSTMs},'' in \emph{Proceedings of the
  International Conference on Machine Learning}, 2015.

\bibitem{mahmood2019}
N.~Mahmood, N.~Ghorbani, N.~Troje, G.~Pons-Moll, and M.~Black, ``{AMASS:
  Archive of Motion Capture as Surface Shapes},'' in \emph{Proceedings of the
  International Conference on Computer Vision}, 2019.

\bibitem{krizhevsky2012imagenet}
A.~Krizhevsky, I.~Sutskever, and G.~E. Hinton, ``Imagenet classification with
  deep convolutional neural networks,'' \emph{Proceedings of the Advances in
  Neural Information Processing Systems}, 2012.

\bibitem{vgg}
K.~Simonyan and A.~Zisserman, ``Very deep convolutional networks for
  large-scale image recognition,'' \emph{arXiv preprint arXiv:1409.1556}, 2014.

\bibitem{iandola2016squeezenet}
F.~N. Iandola, S.~Han, M.~W. Moskewicz, K.~Ashraf, W.~J. Dally, and K.~Keutzer,
  ``Squeezenet: Alex{N}et-level accuracy with 50x fewer parameters and $<$ 0.5
  mb model size,'' \emph{arXiv preprint arXiv:1602.07360}, 2016.

\bibitem{he2016deep}
K.~He, X.~Zhang, S.~Ren, and J.~Sun, ``Deep residual learning for image
  recognition,'' in \emph{Proceedings of the Conference on Computer Vision and
  Pattern Recognition}, 2016.

\bibitem{ji20123d}
S.~Ji, W.~Xu, M.~Yang, and K.~Yu, ``3{D} convolutional neural networks for
  human action recognition,'' \emph{IEEE Transactions on Pattern Analysis and
  Machine Intelligence}, vol.~35, no.~1, pp. 221--231, 2012.

\bibitem{tran2015learning}
D.~Tran, L.~Bourdev, R.~Fergus, L.~Torresani, and M.~Paluri, ``Learning
  spatiotemporal features with 3{D} convolutional networks,'' in
  \emph{Proceedings of the International Conference on Computer Vision}, 2015.

\bibitem{tran2018closer}
D.~Tran, H.~Wang, L.~Torresani, J.~Ray, Y.~LeCun, and M.~Paluri, ``A closer
  look at spatiotemporal convolutions for action recognition,'' in
  \emph{Proceedings of the Conference on Computer Vision and Pattern
  Recognition}, 2018.

\bibitem{feichtenhofer2020x3d}
C.~Feichtenhofer, ``X3{D}: Expanding architectures for efficient video
  recognition,'' in \emph{Proceedings of the Conference on Computer Vision and
  Pattern Recognition}, 2020.

\bibitem{hubel1959receptive}
D.~H. Hubel and T.~N. Wiesel, ``Receptive fields of single neurones in the
  cat's striate cortex,'' \emph{The Journal of physiology}, vol. 148, no.~3,
  pp. 574--591, 1959.

\bibitem{goodale1992separate}
M.~A. Goodale and A.~D. Milner, ``Separate visual pathways for perception and
  action,'' \emph{Trends in Neurosciences}, vol.~15, no.~1, pp. 20--25, 1992.

\bibitem{simonyan2014two}
K.~Simonyan and A.~Zisserman, ``Two-stream convolutional networks for action
  recognition in videos,'' in \emph{Proceedings of the Conference on Advances
  in Neural Information Processing Systems}, 2014.

\bibitem{feichtenhofer2017spatiotemporal}
C.~Feichtenhofer, A.~Pinz, and R.~P. Wildes, ``Spatiotemporal multiplier
  networks for video action recognition,'' in \emph{Proceedings of the
  Conference on Computer Vision and Pattern Recognition}, 2017.

\bibitem{feichtenhofer2019slowfast}
C.~Feichtenhofer, H.~Fan, J.~Malik, and K.~He, ``Slow{F}ast networks for video
  recognition,'' in \emph{Proceedings of the International Conference on
  Computer Vision}, 2019.

\bibitem{wang2018non}
X.~Wang, R.~Girshick, A.~Gupta, and K.~He, ``Non-local {N}eural {N}etworks,''
  in \emph{Proceedings of the Conference on Computer Vision and Pattern
  Recognition}, 2018.

\bibitem{fan2021multiscale}
H.~Fan, B.~Xiong, K.~Mangalam, Y.~Li, Z.~Yan, J.~Malik, and C.~Feichtenhofer,
  ``Multiscale vision transformers,'' in \emph{Proceedings of the International
  Conference on Computer Vision}, 2021.

\bibitem{hadji2018new}
I.~Hadji and R.~P. Wildes, ``A new large scale dynamic texture dataset with
  application to convnet understanding,'' in \emph{Proceedings of the European
  Conference on Computer Vision}, 2018.

\bibitem{ghodrati2018video}
A.~Ghodrati, E.~Gavves, and C.~G.~M. Snoek, ``Video time: Properties, encoders
  and evaluation,'' in \emph{British Machine Vision Conference}, 2018.

\bibitem{manttari2020interpreting}
J.~Manttari, S.~Broom{\'e}, J.~Folkesson, and H.~Kjellstrom, ``Interpreting
  video features: {A} comparison of 3{D} convolutional networks and
  convolutional {L}{S}{T}{M} networks,'' in \emph{Proceedings of the Asian
  Conference on Computer Vision}, 2020.

\bibitem{sevilla2018integration}
L.~Sevilla-Lara, Y.~Liao, F.~G{\"u}ney, V.~Jampani, A.~Geiger, and M.~J. Black,
  ``On the integration of optical flow and action recognition,'' in
  \emph{Proceedings of the German Conference on Pattern Recognition}.\hskip 1em
  plus 0.5em minus 0.4em\relax Springer, 2018.

\bibitem{matt2022}
M.~Kowal, M.~Siam, A.~Islam, N.~D.~B. Bruce, R.~P. Wildes, and K.~G. Derpanis,
  ``A deeper dive into what deep spatiotemporal networks encode: {Quantifying}
  static vs. dynamic information,'' in \emph{Proceedings of the Conference on
  Computer Vision and Pattern Recognition}, 2022.

\bibitem{miller1956magical}
G.~A. Miller, ``The magical number seven, plus or minus two: Some limits on our
  capacity for processing information.'' \emph{Psychological review}, vol.~63,
  no.~2, p.~81, 1956.

\bibitem{thoma2017analysis}
M.~Thoma, ``Analysis and optimization of convolutional neural network
  architectures,'' Master's thesis, University of the State of
  Baden-Wuerttemberg, 2017.

\bibitem{dittrich1993action}
W.~H. Dittrich, ``Action categories and the perception of biological motion,''
  \emph{Perception}, vol.~22, no.~1, pp. 15--22, 1993.

\bibitem{troje2002decomposing}
N.~F. Troje, ``Decomposing biological motion: A framework for analysis and
  synthesis of human gait patterns,'' \emph{Journal of Vision}, vol.~2, no.~5,
  pp. 2--2, 2002.

\bibitem{adam}
D.~P. Kingma and J.~Ba, ``Adam: A method for stochastic optimization,''
  \emph{arXiv preprint arXiv:1412.6980}, 2014.

\end{thebibliography}

\end{document}